# Predicting vacant parking space availability zone-wisely: a graph based spatio-temporal prediction approach

Yajing Feng, Qian Hu, and Zhenzhou Tang

*Abstract*—Vacant parking space (VPS) prediction is one of the key issues of intelligent parking guidance systems. Accurately predicting VPS information plays a crucial role in intelligent parking guidance systems, which can help drivers find parking space quickly, reducing unnecessary waste of time and excessive environmental pollution. Through the simple analysis of historical data, we found that there not only exists a obvious temporal correlation in each parking lot, but also a clear spatial correlation between different parking lots. In view of this, this paper proposed a graph data-based model ST-GBGRU (Spatial-Temporal Graph Based Gated Recurrent Unit), the number of VPSs can be predicted both in short-term (i.e., within 30 min) and in long-term (i.e., over 30min). On the one hand, the temporal correlation of historical VPS data is extracted by GRU, on the other hand, the spatial correlation of historical VPS data is extracted by GCN inside GRU. Two prediction methods, namely direct prediction and iterative prediction, are combined with the proposed model. Finally, the prediction model is applied to predict the number VPSs of 8 public parking lots in Santa Monica. The results show that in the short-term and long-term prediction tasks, ST-GBGRU model can achieve high accuracy and have good application prospects.

*Index Terms*—Vacant parking space availability Deep Learning GCN GRU

## I. INTRODUCTION

Parking has become a common problem in all kinds of cities during the peak traffic flow in urban centers. Restricted by the city's land planning, parking resources are limited and the parking utilization rate within the adjacent area is seriously unevenly distributed. Parking brings traffic congestion, environmental pollution and other problems to the existing production, life, environmental security caused great impact [1, 2, 3, 4]. And in the Internet environment, the use of parking space predicting technology to accurately predict the number of VPS is of great importance to alleviate the urban parking problem.

VPS prediction is an important research topic in the field of time series data mining, and it has been studied for a long time. Earlier forecasting methods include historical average model (HA), time series models, etc. HA [5] is a combination of data from two adjacent time periods by adjusting the ratio and then combining it with the least squares method for modeling, which has the advantage of being computationally simple, but the limitation of the method is that it cannot be accurately predicted for random temporal events and some unanticipated events. Common time series models roughly include Moving Average Model (MA) [6], Auto-Regressive Model (AR) [7],

Auto-Regressive Moving Average Model (ARMA) [8], Auto-Regressive Integrated Moving Average Model (ARIMA) [9].

With the continuous updating and iteration of data mining technology, machine learning models started to replace other methods for prediction, such as Support Vector Regression model (SVR), Extreme Gradient Boosting model (XGB), Gradient Boosting Decision Tree model (GBDT), and K-Nearest Neighbor model (KNN). J. Fan [10] combined the SVR model with a Fruit Fly Optimization Algorithm (FOA) to propose a new parking space prediction model for predicting the number of VPS in a single parking lot. D. Chuan's study [11] showed that the XGB model outperformed the Random Forest model (RF) in terms of prediction performance and efficiency. X. Ye [12] proposed an algorithm for VPS prediction based on GBDT and Wavelet Neural Network (WNN). On this basis, an improved WNN algorithm combining Wavelet Analysis (WA) decomposition and Particle Swarm Optimization (PSO) was proposed. Habtemuchael et al. [13] developed an improved KNN model that can be used for traffic flow prediction at a single point in a short period of time.

Along with the development of neural networks, among which Recurrent Neural Networks (RNN) are widely used in time series prediction tasks, typically Long Short-Term Memory (LSTM), Gated Recurrent Unit (GRU). Y. Du [14] used an RNN model for vehicle speed prediction. J. Fan [15] built the LSTM model and proposed two multi-step prediction methods to predict the number of VPS in a single parking lot for 5-60 minutes. C. Zeng [16] proposed a stacked GRU-LSTM model for VPS prediction.

While RNNs are able to capture the correlation of sequences in the temporal dimension, they cannot consider the correlation of nodes in the spatial dimension. Deep-learning-based prediction model for Spatial-Temporal data (Deep_ST) [17] is the first model to capture grid spatial dependencies using CNN. Based on this, Deep Spatio-Temporal Residual Network (ST_ResNet) [18] adopted a Residual Network (ResNet) framework, considered three different time trend to exploit spatiotemporal relationships, and finally integrates external information (i.e. weather) to predict traffic flow. And reference [19] proposed a DeepSTN+ model, which improved the internal structure of the residual unit and can more accurately express the spatiotemporal characteristics of the sequence. ST-3DNet [20] introduces 3D convolution to capture the correlation of traffic data in spatial and temporal dimensions, taking into account two characteristics of traffic data, i.e., short-term and long-term.

Although these aforementioned models are able to capture spatial information using CNNs, the input of such models is restricted to standard grid data and cannot be used for graph structure prediction problems. In recent years, scholars have started to use GCNs to model correlations in the spatial dimension of graph structure data. Fang et al. [21] proposed a Global Spatiotemporal Network (GSTNet) composed of multi-layered spatiotemporal modules to predict traffic flow. Reference [22] proposed the Attention Based Spatial-Temporal Graph Convolutional Network (ASTGCN) model, which introduced an attention mechanism into GCN and captured the spatial dependencies through GCN. Li et al. [23] proposed a Diffusion Convolutional Recurrent Neural Network (DCRNN) that uses diffusion convolution and encoder-decoder architecture to capture spatial and temporal correlation. Reference [24] used a Temporal Multi-Graph Convolutional Network (T-MGCN) to model the two semantic associations that exist between roads, using RNN to learn temporal correlations. Wu [25] uses diffusion convolution to capture spatial dependencies and dilated causal convolution as the temporal convolution layer to capture temporal dependencies.

Most of the existing graph-based spatiotemporal prediction methods extract spatiotemporal features separately rather than simultaneously. Therefore, proposing a graph-based model capable of simultaneously extracting spatiotemporal features is of great significance for predicting the number of VPSs zone-wisely. Motivated by the above motivations, this paper proposes a deep learning model combining GCN and GRU to predict the number of VPSs zone-wisely in short-term and long-term. The main contributions of this paper are summarized as follows:

- This paper proposes a new ST-GBGRU model to predict the number of VPSs of all parking lots. The model integrates GCN and GRU, and simultaneously extracts the spatiotemporal features of parking lots. The model is more suitable for VPS prediction in real scenes, and can achieve good prediction performance.
- In this paper, the ST-GBGRU model is combined with two prediction methods, namely short-term prediction and long-term prediction, to significantly improve the prediction accuracy.

The rest of this paper is organized as follows: Section II introduced the proposed ST-GBGRU model in detail. In Section III, the comparative experimental results and analysis are carried out. At last, Section IV is the conclusion of the paper.

## II. METHODOLOGY

This section describes ST-GBGRU framework in detail, and puts forward the method of predicting the number of VPSs.

### A. Data description & preprocessing

The dataset come from the real parking data of 8 parking lots (St1-St8) in Santa Monica, California, USA [26]. The data collection frequency of each parking lot is 5 minutes, and each parking lot has 9108 recorded data. The location of the parking lots is: longitude range [-118.499378, -118.49372], latitude range [34.019575, 34.01289]. The starting and ending time of data collection is from 05/11/2018 7:00 to 06/11/2018 23:55.

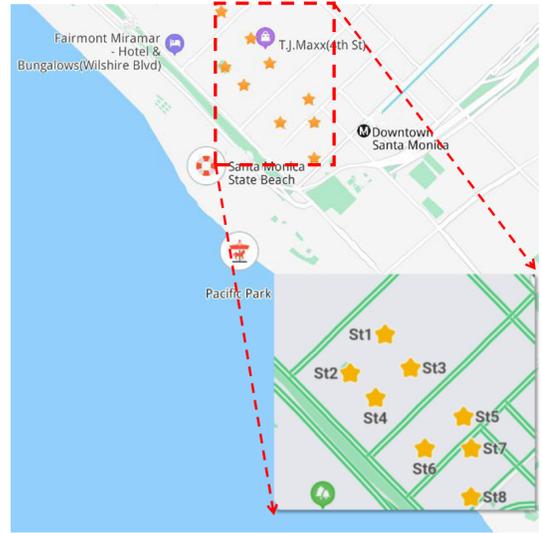

**Fig. 1** Distribution of the parking lots.

The preprocessing steps of the dataset are as follows: (1) Build a graph based on the geographic distance of the parking lot; (2) Use the Min-Max normalization method to scale all the data to the range [0,1]; (3) Use the sliding window method to intercept the dataset, and reshape the data into samples of a specified length for training, where the training data is in the form of $[X_1, X_2, X_3, ..., X_n]$, and $X$ represents the input, $n$ represents the length of the input sequence.

The graph of the parking lot is built according to the geographical distance between parking lots. We describe the spatial dependence between parking lots by constructing an adjacency matrix. The calculation process of the adjacency matrix is as follows: (1) Using the latitude and longitude of each parking lot, calculate the distance between all parking lots according to formula (1); (2) According to the formula (2), the adjacency matrix is calculated using geographic distance. The geographical distance is calculated as follows:

$$d_{ij} = 2R. \arcsin \sqrt{\sin^2(a) + \cos(lat_i)\cos(lat_j)\sin^2(b)}, \tag{1}$$

$$A_{ij} = \left\{ \begin{array}{ll} d_{ij}, d_{ij} \leq \varepsilon \\ 0, \quad d_{ij} > \varepsilon \end{array} \right. . \tag{2}$$

Here $a = \left( \frac{lat_i - lat_j}{2} \right)$, $b = \frac{long_i - long_j}{2}$, $d_{ij}$ represents the geographic distance from the parking lot $i$ to the parking lot $j$; $R$ is the radius of the earth; $lat_i, long_i$ is the longitude and latitude of the parking lot $i$; and $\varepsilon$ is the address distance threshold. After many experiments, set the value of $\varepsilon$ to 0.35.

In our previous work [27, 28], we have fully analyzed the temporal and spatial correlation of parking lot. In view of this, the specific analysis is not carried out in this paper, and we simply show the results of the spatiotemporal analysis. Fig. 2a and 2b are the temporal correlation diagram of parking lot St1,

Fig. 2c shows the spatial correlation of parking lots St1-St8. It can be seen from the figure that there is obvious temporal correlation in each parking lot and spatial correlation between different parking lots. Next, the prediction model is introduced in detail.

### B. Prediction Model

*1) GCN:* Gragh Convolutional Networks (GCN) [29, 30, 31] is a variant of CNN, which is a natural extension of CNN in graph domain. Graph data is irregular and does not have local translation invariance like images, so traditional CNN cannot perform convolution operations on graph data. In order to generalize and apply traditional CNN operations to graph data, scholars propose to use GCN to process graph data in deep learning. The basic goal of GCN is to extract the spatial features from the graph, among which the realization of spatial domain [32] and the spectral domain [31] are the two mainstream implementations. The GCN operation used in this paper is the latter, whose basic idea is similar to that of CNN (i.e. weighted average operation of pixels within a certain range).

Fig. 3a is a detailed introduction to GCN. GCN captures the spatial dependencies of a graph through messages passing between graph nodes. In order to define GCN, according to the theory proposed in [33], for graph $G = (V, E, A)$, input $X$ and output $Y$, the data processing method for GCN is defined as: $f(X, A) = Y$. Here $V \in \mathbf{R}^N$ represents the node set, that is, the set composed of all nodes in the graph, $N$ represents the number of nodes; $E \in \mathbf{R}^{N \times N}$ represents the set of all edges; $A$ represents the adjacency matrix, $A \in R^{N \times N}$; and the element $A_{ij}$ of $A$ in the matrix represents the connection between the node $v_i$ and the $v_j$ in the graph $G$. The feature propagation between layers of GCN can be expressed as follows:

$$H^{l+1} = \sigma(\tilde{D}^{-\frac{1}{2}}\hat{A}\hat{D}^{\frac{1}{2}}H^l W^l). \tag{3}$$

Here $\tilde{A} = A + I$ is the matrix with add self-connections, $I$ is a $N \times N$ identity matrix; $\tilde{D}$ is the degree matrix in the form of diagonal matrix , $\tilde{D}_{ij} = \sum_j A_{ij}$; $H^l \in R^{N \times C}$ represents the output features of $l$-th layer, when $l = 1$, $H^{l-1} = H^0 = X$, $C$ is the dimension of output features; $\sigma(.)$ represents the activation function; $W^l$ represents the learnable parameters of the $l$-th layer.

In this paper, we chose a 2-layer GCN [34] to capture spatial dependence, the data processing method for GCN can be expressed as

$$f(X, A) = \sigma(\hat{A}\text{ReLU}(\hat{A}XW_0)W_1). \tag{4}$$

Where $\hat{A} = \tilde{D}^{-\frac{1}{2}}A\tilde{D}^{-\frac{1}{2}}$, ReLU(.) represents Rectified Linear Unit, $W_0, W_1$ are weights.

*2) GRU:* Gated Recurrent Unit (GRU) [35] was proposed to address the computationally overly complex problem of Long Short-Term Memory (LSTM) [36]. The principle of GRU is very similar to that of LSTM, i.e. the gate mechanism is used to control input, memory and other information to make predictions at the current time step. The internal structure of GRU is shown in Fig. 3b.

As can be seen from Fig. 3b, the GRU consists of two gates, namely the reset gate $z_t$ and the update gate $r_t$. Intuitively, the reset gate mainly determines how much past information to forget, and the update gate helps the model decide how much past information to pass. The expression for GRU is as follows:

$$
\begin{aligned}
r_t &= \sigma(x_t \mathbf{w}_{xr} + h_{t-1} \mathbf{w}_{hr} + b_r), \\
z_t &= \sigma(x_t \mathbf{w}_{xz} + h_{t-1} \mathbf{w}_{hz} + b_z), \\
\tilde{h}_t &= \tanh(x_t \mathbf{w}_{xh} + \mathbf{w}_{hh}(r_t \odot h_{t-1})), \\
h_t &= z_t \odot h_{t-1} + (1 - z_t) \odot \tilde{h}_t.
\end{aligned}
\tag{5}
$$

Where, $x_t$ represents the input of the current time step $t$, $r_t$ is the update gate, $z_t$ is the reset gate, $h_{t-1}$ carries the information of the previous time step $t - 1$, $h_t$ keeps information from the current memory cell and passes it to the next memory cell, $\tanh(.)$ represents the hyperbolic tangent function, $\mathbf{w}_{xr}, \mathbf{w}_{xz}, \mathbf{w}_{xh}, \mathbf{w}_{hr}, \mathbf{w}_{hz}, \mathbf{w}_{hh}$ represents weights, and $b_r, b_z$ are bias.

*3) ST-GBGRU:* Reference [37] has proposed a model of stacking GCN+GRU. This method ignores the spatial feature extraction of the output of the hidden layer, however the output of the hidden layer has an influence on the accuracy of the result. If the features generated by the hidden layer are not spatially extracted, the spatial and temporal extraction features are insufficient, thus affecting the prediction accuracy. Reference [38] uses deep learning method to estimate uncertainty, and put forward a GCN-GRU-opinion model. GCN-GRU-opinion model is the GCN-GRU-opinion model integrates an GRU-based opinion model to model the time series relational dependencies between beliefs and uncertainties based on dynamic structural dependencies. Although the spatiotemporal features are extracted simultaneously, the model is complex and does not predict the number of VPS. In this perspective, this article proposes a deep learning based prediction model. The model integrates GCN and GRU, which can fully extract the spatial features of the hidden layer output and achieve good prediction performance.

This subsection will introduce the proposed ST-GBGRU model in detail. First, the detail structure and training process of the ST-GBGRU model are introduced in Fig. 5. The model input consists of two parts, a graph of parking lot and a time series of VPS. The basic structure of the ST-GBGRU model is composed of two models, GCN and GRU. The role of the GCN network is to extract the spatial characteristics of VPS, and the role of the GRU network is to extract the temporal characteristics of VPS. We merge GCN and GRU models and use GCN to extract spatial features from input and hidden layers. The expression for ST-GBGRU is as follows:

$$
\begin{aligned}
z_t &= \sigma(\mathbf{W}_{xz *\mathcal{G}} X_t + \mathbf{W}_{hz *\mathcal{G}} H_{t-1} + b_z), \\
r_t &= \sigma(\mathbf{W}_{xr *\mathcal{G}} X_t + \mathbf{W}_{hr *\mathcal{G}} H_{t-1} + b_r), \\
\tilde{H}_t &= \tanh(\mathbf{W}_{xh *\mathcal{G}} X_t + \mathbf{W}_{hh *\mathcal{G}}(r_t \odot H_{t-1})), \\
H_t &= z_t \odot H_{t-1} + (1 - z_t) \odot \tilde{H}.
\end{aligned}
\tag{6}
$$

Where $*\mathcal{G}$ represents the GCN operation on the graph, $X_t$ represents the input of the current time step $t$, $r_t$ is the update gate, $z_t$ is the reset gate, $h_{t-1}$ carries the information of the previous time step $t - 1$, $H_t$ keeps information from

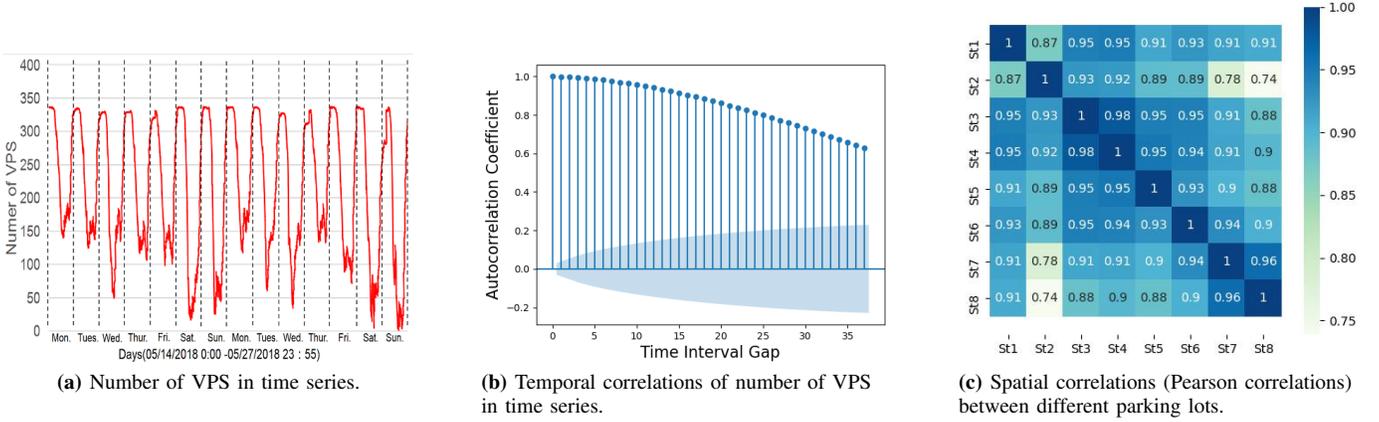

**(a)** Number of VPS in time series.

**(b)** Temporal correlations of number of VPS in time series.

**(c)** Spatial correlations (Pearson correlations) between different parking lots.

**Fig. 2** The spatial and temporal dynamics of parking space.

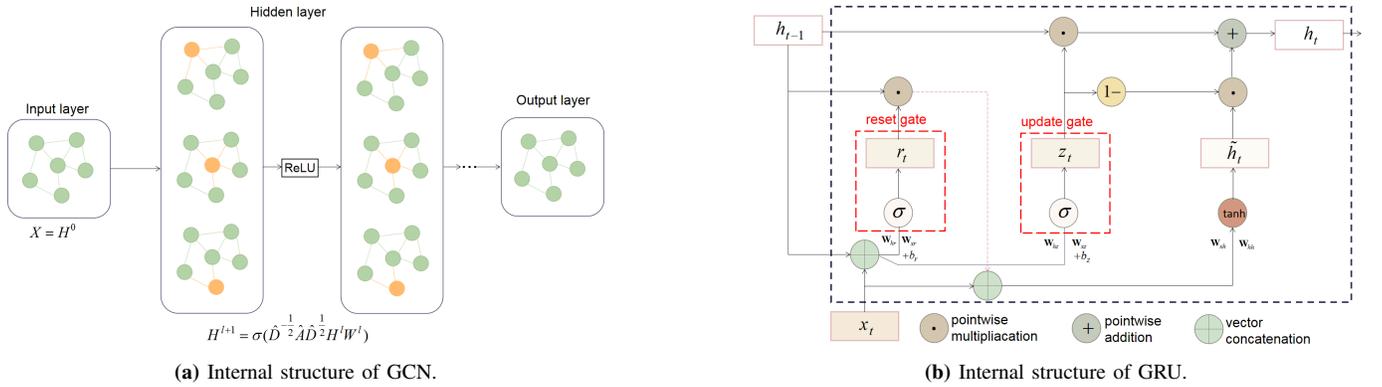

**(a)** Internal structure of GCN.

**(b)** Internal structure of GRU.

**Fig. 3** Detailed introduction of GCN and GRU.

the current memory cell and passes it to the next memory cell, $\tanh(.)$ represents the hyperbolic tangent function, $\mathbf{W}_{xr}, \mathbf{W}_{xz}, \mathbf{W}_{xh}, \mathbf{W}_{hr}, \mathbf{W}_{hz}, \mathbf{W}_{hh}$ represents weights, and $b_r, b_z$ are bias.

### C. Prediction Methods

The historical VPS sequence $[X_{t_c-(m-1)\delta}, X_{t_c-(m-2)\delta}, \ldots, X_{t_c}]$ is used to predict $\hat{Y}_{t_c+h\delta}$ (i.e. the number of VPSs at time $t_c + h\delta$). According to the length of the prediction time step, the prediction method can be divided into short-term and long-term prediction, in which the prediction within 30 min (i.e. $h\delta \leq 30$ min) is called short-term prediction, and the prediction over 30 min (i.e. $h\delta > 30$ min) is called long-term prediction, as shown in in Fig. 6.

For short-term and long-term prediction, two forecasting methods are proposed, namely single-step direct prediction and multi-step iterative prediction. To be exact, as shown in the first part of Fig. 6, in single-step direct prediction, $m$ historical observation data are directly used to predict $\hat{Y}_{t_c+h\delta}$. However, in the multi-step iterative prediction method, as shown in the second part of Fig. 6, first use the single-step direct prediction method with $m$ historical observations $[X_{t_c-(m-1)\delta}, X_{t_c-(m-2)\delta}, \ldots, X_{t_c}]$ to predict $\hat{Y}_{t_c+\delta}$, then

$[X_{t_c-(m-2)\delta}, \ldots, X_{t_c}, \hat{Y}_{t_c+\delta}]$ is used to predict $\hat{Y}_{t_c+2\delta}$. And so on, the final prediction result $\hat{Y}_{t_c+h\delta}$ is obtained.

### D. Evaluation indicators and experimental setup

*1) Evaluation indicators:* This paper adopts Root Mean Square Error (RMSE), Mean Absolute Percentage Error (MAPE) and Mean Absolute Error (MAE) to evaluate the performance of the model. Concretely, MAE measures the average absolute error between the actual values and predicted value. MAE can precisely reflect forecast errors for it avoids the problem of positive and negative errors interacting with each other. The expression of MAE is as follows:

$$\text{MAE} = \frac{1}{T}\sum_{t=1}^{T}\left|y_t^i - \hat{y}_t^i\right|, \tag{7}$$

where $y_t^i$ and $\hat{y}_t^i$ are the actual number of VPSs and predicted numbers of VPSs of $i$-th parking lot at time $t$, and $T$ represent the time step.

MAPE uses percentages to measure the relative magnitude of the deviation and its expression is as follows:

$$\text{MAPE} = \frac{100\%}{T}\sum_{t=1}^{T}\frac{\left|y_t^i - \hat{y}_t^i\right|}{y_t^i}. \tag{8}$$

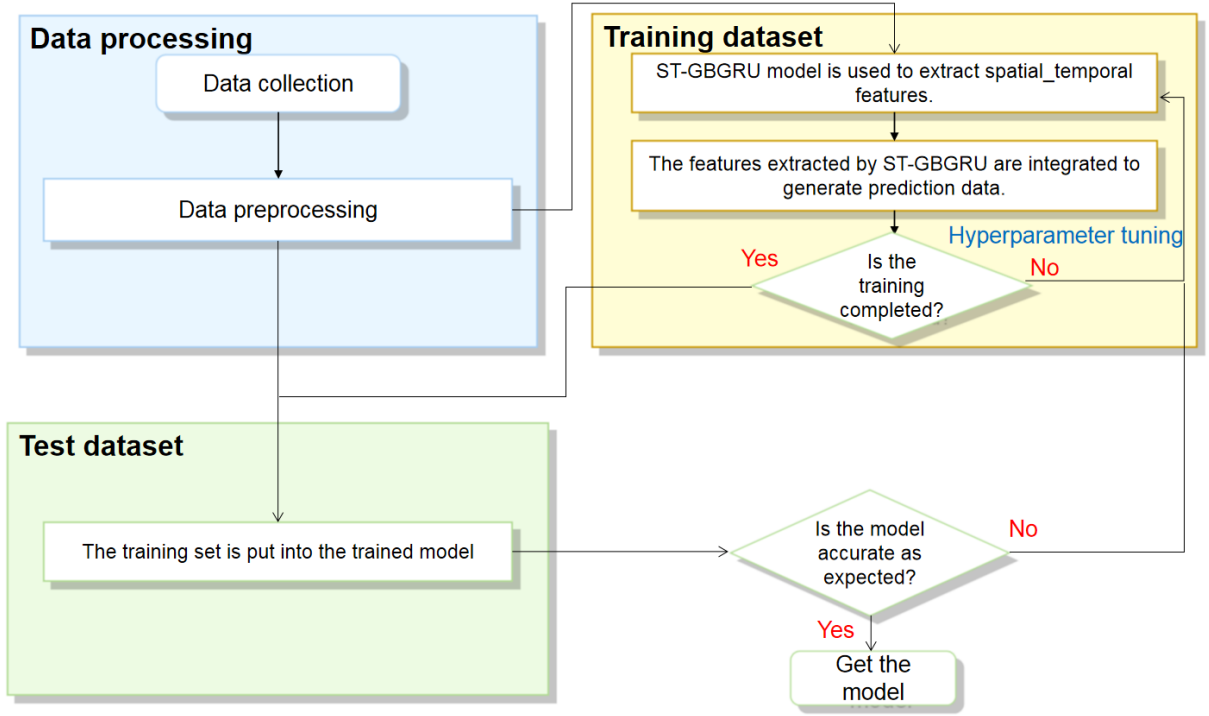

**Fig. 4** Flow prediction process of ST-GBGRU model.

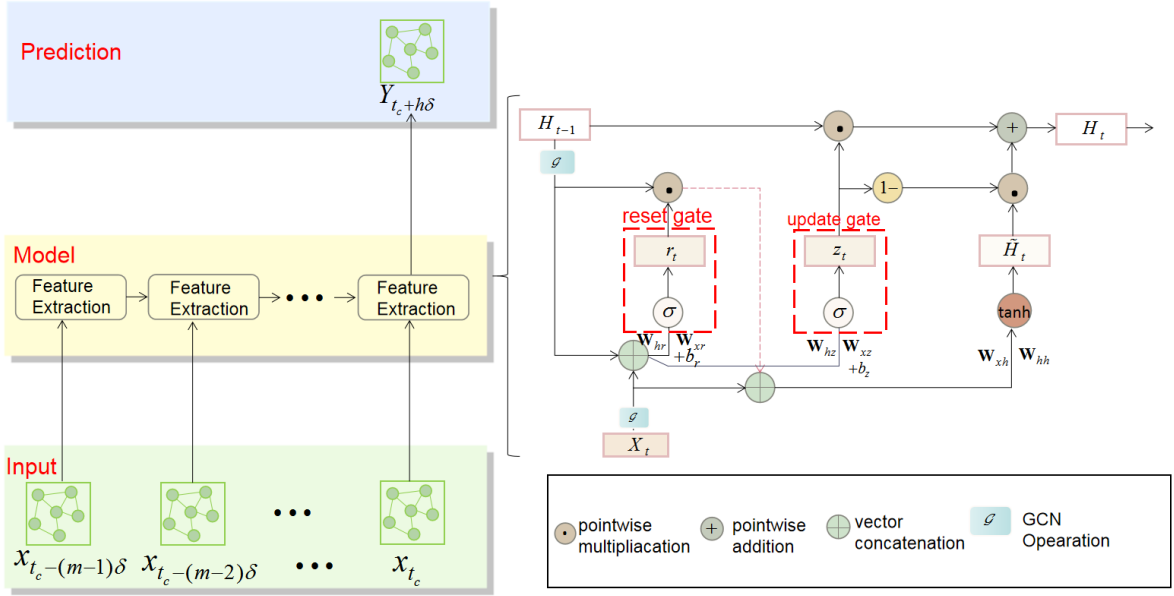

**Fig. 5** Overview of ST-GBGRU model.

The disadvantage of MAPE is that it is not computable (a division by zero error occurs) when there are cases where the true value is zero. However, in parking lot St3 and St5, this non-computable MPAE situation occurs. Therefore, in order to avoid this situation, use SMAPE (Symmetric MAPE) instead of MAPE. The formula of SMAPE is as follows:

$$\text{SMAPE} = \frac{100\%}{T} \sum_{t=1}^{T} \frac{\left| y_t^i - \hat{y}_t^i \right|}{\left( \left| \hat{y}_t^i \right| + \left| y_t^i \right| \right) / 2}. \quad (9)$$

RMSE reflects the degree of deviation between the actual value and the predicted value, and its expression is as follows:

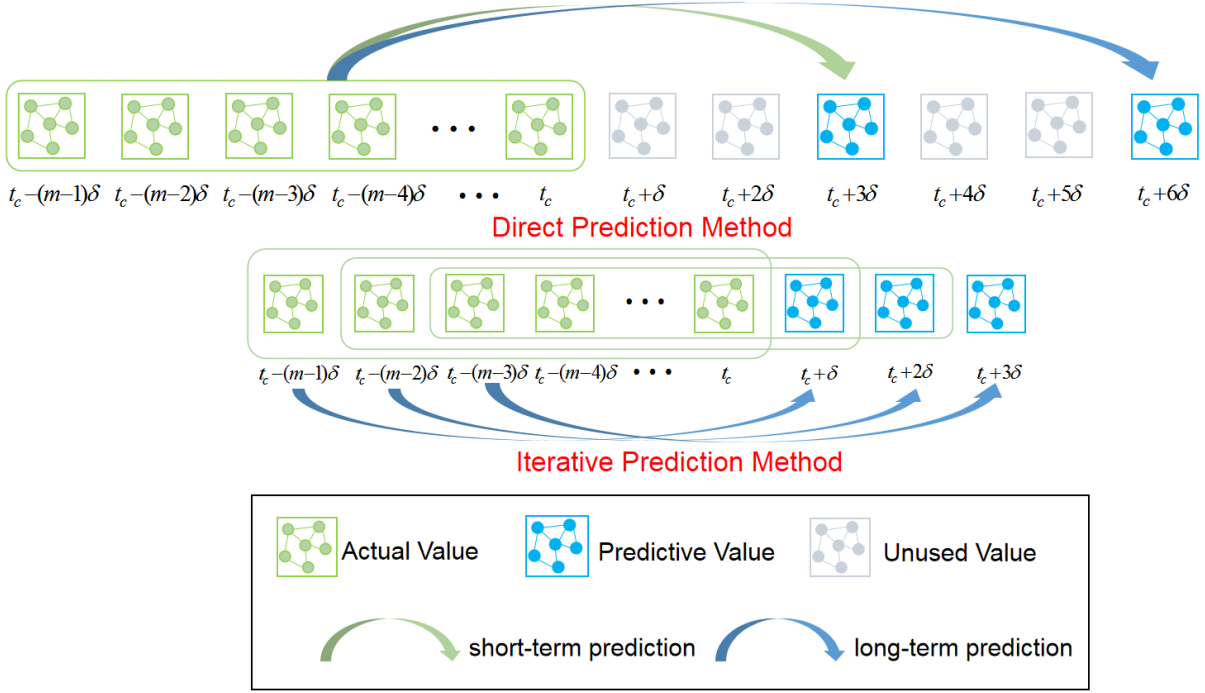

**Fig. 6** Four prediction methods, i.e. short-term direct prediction, long-term direct prediction, short-term iterative prediction and short-term direct prediction methods. The green arrows represent the short-term predictions and the blue arrows represent long-term predictions.

$$\text{RMSE} = \sqrt{\frac{1}{T} \sum_{t=1}^{T} (y_t^i - \hat{y}_t^i)^2}. \qquad (10)$$

Generally speaking, if the values of the above three indicators are smaller, the prediction performance of the prediction model is better.

*2) Experimental setup:* The ConvLSTM-DCN model [27] and the GCN+GRU model [37] (i.e. stacked GCN and GRU models) are used as baseline models to evaluate the performance of the ST-GBGRU model.

Experiments were conducted to predict the number of VPSs in 5-/15-/30-/45-/60- minutes. To avoid the contingency of experimental results due to the random parameter initialization of the neural network, each prediction task was independently performed 50 times, and the average values were taken. The model hyperparameters of ST-GBGRU are as follows: the epoch size is 500, the batch size is 32, the length of the time series ($m$) is 12, and the learning rate is 0.001. The ratio of training set and test set is 4:1. The settings of the ConvLSTM-DCN model hyperparameters can be found in [27]. The hyperparameter settings for the GCN+GRU model are the same as that of the ST-GBGRU model. The above models are trianed using the Adam optimizer [39], and the hyperparameters of the models are selected using the grid search method. The experiments were conducted on a 64-bit server, the configuration of the server is shown in Table I.

**TABLE I Configuration information**

| Attribute | Configuration |
|---|---|
| Operating System | Windows 10 |
| CPU | Intel® Xeon E5-2620, 2.10GHz |
| GPU | NVIDIA GeForce GTX 1080 Ti |
| RAM | 64GB |
| Pycharm | JetBrains Pycharm Community Edition 2019.2.4 |
| Framework | TensorFlow |

*E. Results and analysis*

To evaluate the prediction performance of the ST-GBGRU model proposed in this paper. Two series of experiments, direct single-step prediction and iterative multi-step prediction, are conducted on 8 parking lots in Santa Monica. It should be noted that, unlike our previous work [27], [28], the distance between the parking lot St9 and the other 8 parking lots is far away, which will affect the extraction of spatial features, thus affecting the prediction results. Therefore, in this paper, St1-St8 are used to predict the number of VPSs. Table II shows the overall predictive performance of different models in predicting the number of VPS in the next 5-60 minutes on the Santa Monica dataset, where the best results are marked in bold red. Lower values of the three evaluation indicators represent better prediction performance of the model. Besides, Fig. 7 and Fig. 8 are the comparison results between the predicted value and the actual value at 5, 15, 30, 45, 60 minutes. The $x$-axis represents the time interval, $y$-axis represents the number of VPSs, the red curve represents the predicted value, and the black curve represents the actual value. The closer the curve of predicted

**TABLE II Performance evaluations in terms of MAE, RMSE and MAPE/SMAPE.**

| Indicators | Parking lot | 5min | 15min | 30min | 45min | 60min | 5min | 15min | 30min | 45min | 60min |
|---|---|---|---|---|---|---|---|---|---|---|---|
| | | ConvLSTM-DCN,Direct | | | | | ConvLSTM-DCN,Iterative | | | | |
| | St1 | 3.40 | 7.18 | 10.95 | 14.08 | 16.65 | — | 6.66 | 10.60 | 16.25 | 20.91 |
| | St2 | 4.01 | 6.86 | 11.04 | 14.33 | 16.95 | — | 7.78 | 13.40 | 22.23 | 26.54 |
| | St3 | 3.70 | **5.62** | **8.62** | 10.99 | 12.44 | — | 6.61 | 11.04 | 17.50 | 18.79 |
| | St4 | 4.00 | **6.15** | **9.27** | 12.03 | 14.46 | — | 7.16 | 12.84 | 19.24 | 24.67 |
| | St5 | 5.23 | **6.77** | 10.50 | 14.06 | 17.25 | — | 8.99 | 14.16 | 21.81 | 25.77 |
| | St6 | 5.46 | **9.59** | 15.85 | 21.12 | 24.38 | — | 10.93 | 20.29 | 31.22 | 37.56 |
| | St7 | 5.38 | 7.92 | 12.02 | 16.13 | 19.30 | — | 9.36 | 16.12 | 22.29 | 25.32 |
| | St8 | 7.46 | 9.16 | 13.31 | 17.10 | 20.50 | — | 10.05 | 16.17 | 26.55 | 32.56 |
| | | GCN+GRU,Direct | | | | | GCN+GRU,Iterative | | | | |
| MAE | St1 | 13.42 | 16.05 | 16.39 | 17.17 | 18.73 | — | 16.58 | 27.96 | 28.97 | 45.57 |
| | St2 | 17.80 | 20.41 | 23.47 | 23.53 | 25.67 | — | 23.39 | 29.43 | 43.06 | 57.49 |
| | St3 | 11.41 | 13.41 | 14.31 | 14.39 | 15.06 | — | 15.51 | 17.66 | 21.45 | 40.10 |
| | St4 | 12.81 | 14.28 | 14.28 | 18.03 | 19.62 | — | 18.45 | 26.12 | 33.36 | 64.88 |
| | St5 | 24.96 | 27.75 | 30.68 | 34.25 | 34.43 | — | 32.48 | 33.56 | 55.83 | 70.60 |
| | St6 | 23.58 | 24.61 | 26.59 | 32.11 | 33.86 | — | 26.98 | 40.35 | 45.51 | 75.37 |
| | St7 | 14.43 | 13.61 | 22.06 | 23.31 | 26.92 | — | 26.07 | 31.21 | 42.71 | 98.39 |
| | St8 | 26.71 | 27.24 | 29.33 | 29.96 | 32.82 | — | 22.77 | 27.83 | 49.73 | 82.80 |
| | | ST-GBGRU,Direct | | | | | ST-GBGRU,Iterative | | | | |
| | St1 | **3.26** | **6.39** | **6.86** | **7.73** | **9.52** | — | 6.04 | 12.71 | 13.83 | 13.92 |
| | St2 | **3.65** | **5.77** | **7.87** | **10.32** | **12.94** | — | 7.11 | 13.50 | 17.95 | 23.10 |
| | St3 | **3.31** | 9.69 | 9.79 | **10.60** | **11.92** | — | 6.83 | 13.87 | 14.92 | 15.50 |
| | St4 | **3.77** | 9.69 | 9.94 | **11.48** | **13.33** | — | 8.22 | 15.77 | 18.57 | 19.64 |
| | St5 | **3.96** | 9.05 | **10.21** | **11.72** | **13.17** | — | 6.70 | 12.25 | 15.73 | 18.60 |
| | St6 | **5.12** | 10.95 | **12.63** | **15.29** | **17.89** | — | 10.02 | 18.90 | 23.21 | 27.65 |
| | St7 | **4.94** | **7.35** | **9.58** | **11.96** | **14.81** | — | 9.27 | 16.07 | 19.48 | 22.35 |
| | St8 | **4.49** | **8.76** | **11.43** | **14.62** | 18.22 | — | 9.91 | 16.13 | 21.96 | 29.17 |
| | | ConvLSTM-DCN,Direct | | | | | ConvLSTM-DCN,Iterative | | | | |
| | St1 | 4.62 | **7.18** | 10.95 | 14.08 | 16.65 | — | 8.45 | 15.16 | 23.50 | 28.20 |
| | St2 | 5.54 | 9.61 | 15.47 | 20.12 | 24.00 | — | 11.17 | 19.24 | 32.22 | 37.38 |
| | St3 | 5.06 | **7.88** | **12.07** | 15.68 | 18.22 | — | 9.33 | 15.42 | 25.62 | 29.05 |
| | St4 | 5.48 | **8.51** | **12.94** | 17.18 | 21.04 | — | 9.79 | 17.29 | 26.40 | 34.51 |
| | St5 | 14.15 | 19.99 | 27.71 | 33.62 | 38.85 | — | 24.26 | 31.70 | 40.23 | 46.77 |
| | St6 | 9.91 | 16.13 | 23.38 | 28.86 | 32.50 | — | 17.49 | 28.14 | 42.31 | 50.65 |
| | St7 | 7.57 | 11.14 | 16.91 | 23.14 | 28.67 | — | 12.90 | 22.06 | 31.65 | 36.86 |
| | St8 | 10.04 | 12.76 | 18.94 | 24.39 | 30.17 | — | 13.89 | 23.28 | 38.62 | 47.80 |
| | | GCN+GRU,Direct | | | | | GCN+GRU,Iterative | | | | |
| RMSE | St1 | 19.13 | 21.78 | 23.41 | 23.50 | 25.97 | — | 24.40 | 41.74 | 40.32 | 62.00 |
| | St2 | 22.78 | 25.46 | 29.75 | 31.35 | 32.02 | — | 29.20 | 40.82 | 56.34 | 79.39 |
| | St3 | 19.09 | 19.10 | 19.71 | 21.20 | 21.61 | — | 21.41 | 24.12 | 29.76 | 54.94 |
| | St4 | 20.34 | 20.38 | 22.35 | 26.22 | 28.73 | — | 26.00 | 35.91 | 51.11 | 87.19 |
| | St5 | 55.60 | 57.39 | 60.10 | 71.41 | 71.54 | — | 67.81 | 68.59 | 109.46 | 111.30 |
| | St6 | 30.68 | 33.05 | 34.53 | 42.09 | 44.96 | — | 34.59 | 54.34 | 59.25 | 98.22 |
| | St7 | 18.30 | 18.76 | 31.18 | 34.37 | 38.26 | — | 33.45 | 41.80 | 55.84 | 75.78 |
| | St8 | 43.00 | 47.67 | 49.51 | 52.80 | 53.36 | — | 35.33 | 38.29 | 75.09 | 119.47 |
| | | ST-GBGRU,Direct | | | | | ST-GBGRU,Iterative | | | | |
| | St1 | **4.36** | 8.92 | **9.61** | **11.16** | **13.81** | — | 7.98 | 16.07 | 18.45 | 19.48 |
| | St2 | **4.82** | **8.18** | **11.62** | **15.29** | **19.01** | — | 10.09 | 18.65 | 24.19 | 30.67 |
| | St3 | **4.71** | 13.43 | 13.69 | **15.28** | **17.24** | — | 9.18 | 17.12 | 20.28 | 21.15 |
| | St4 | **5.21** | 13.74 | 14.37 | **16.53** | **19.23** | — | 11.08 | 19.55 | 24.77 | 27.22 |
| | St5 | **11.83** | **16.55** | **19.98** | **23.29** | **26.43** | — | 17.28 | 25.99 | 32.40 | 37.08 |
| | St6 | **9.65** | **16.00** | **19.29** | **23.24** | **27.02** | — | 16.28 | 26.60 | 32.81 | 38.00 |
| | St7 | **6.79** | **10.56** | **14.10** | **17.84** | **22.24** | — | 12.45 | 21.38 | 26.77 | 32.20 |
| | St8 | **6.58** | **13.14** | **17.78** | **23.00** | 28.60 | — | 14.53 | 24.03 | 32.99 | 42.96 |
| | | ConvLSTM-DCN,Direct | | | | | ConvLSTM-DCN,Iterative | | | | |
| | St1 | 4.62 | **6.42** | 9.57 | 12.74 | 15.46 | — | 7.16 | 10.80 | 14.73 | 18.51 |
| | St2 | 1.06 | 1.84 | 2.97 | 3.88 | 4.62 | — | 2.12 | 3.66 | 6.18 | 7.07 |
| | St3 | **5.81** | **7.89** | **10.15** | **11.57** | **13.10** | — | 9.97 | 13.85 | 18.06 | 18.82 |
| | St4 | 1.27 | **1.98** | **2.97** | 3.91 | 4.72 | — | 2.28 | 4.00 | 5.84 | 7.49 |
| | St5 | 4.10 | 5.10 | 5.25 | 5.53 | 6.34 | — | 9.24 | 9.85 | 11.97 | 12.07 |
| | St6 | **1.74** | **2.97** | 4.91 | 6.73 | 7.98 | — | 3.48 | 6.39 | 10.47 | 12.05 |
| | St7 | 2.01 | **3.08** | 4.57 | 6.08 | 7.49 | — | 3.54 | 5.92 | 7.50 | 9.23 |
| | St8 | 1.82 | 2.51 | 3.58 | 4.43 | 5.16 | — | 2.64 | 4.09 | 6.06 | 7.34 |
| | | GCN+GRU,Direct | | | | | GCN+GRU,Iterative | | | | |
| MAPE% | St1 | 20.69 | 26.41 | 27.76 | 29.72 | 31.54 | — | 25.94 | 38.74 | 42.27 | 64.82 |
| | St2 | 4.52 | 5.51 | 6.19 | 6.57 | 6.70 | — | 6.41 | 8.27 | 11.65 | 16.96 |
| | St3 | 14.53 | 14.76 | 14.98 | 14.99 | 15.36 | — | 15.76 | 16.63 | 23.81 | 29.17 |
| | St4 | 3.98 | 4.56 | 4.58 | 5.80 | 6.22 | — | 6.15 | 8.78 | 11.47 | 22.74 |
| | St5 | 7.04 | 7.76 | 8.02 | 8.44 | 8.60 | — | 8.60 | 8.90 | 12.90 | 16.55 |
| | St6 | 8.06 | 8.67 | 9.39 | 11.04 | 12.51 | — | 8.81 | 12.32 | 15.83 | 24.08 |
| | St7 | 5.07 | 5.25 | 8.01 | 8.74 | 10.47 | — | 9.57 | 12.32 | 15.42 | 29.47 |
| | St8 | 5.23 | 5.42 | 5.42 | 5.69 | 5.76 | — | 5.19 | 6.80 | 9.69 | 15.16 |
| | | ST-GBGRU,Direct | | | | | ST-GBGRU,Iterative | | | | |
| | St1 | **3.64** | 7.82 | **7.88** | **8.68** | **10.59** | — | 6.49 | 13.82 | 14.97 | 15.72 |
| | St2 | **0.94** | **1.50** | **2.05** | **2.67** | **3.34** | — | 1.86 | 3.50 | 4.65 | 5.82 |
| | St3 | 11.12 | 11.62 | 11.79 | 12.89 | 13.88 | — | 11.30 | 17.19 | 15.19 | 16.03 |
| | St4 | **1.26** | 3.18 | 3.26 | **3.71** | **4.30** | — | 2.77 | 4.91 | 5.94 | 6.26 |
| | St5 | **3.68** | **3.64** | **3.82** | **4.05** | **4.22** | — | 3.07 | 3.89 | 4.57 | 4.99 |
| | St6 | 1.75 | 3.44 | 3.84 | **4.63** | **5.38** | — | 3.03 | 5.79 | 6.81 | 8.01 |
| | St7 | **1.89** | 3.28 | **4.27** | **5.39** | **6.90** | — | 4.12 | 7.69 | 9.41 | 10.56 |
| | St8 | **1.30** | **1.99** | **2.53** | **3.08** | **3.70** | — | 2.38 | 3.42 | 4.61 | 5.52 |

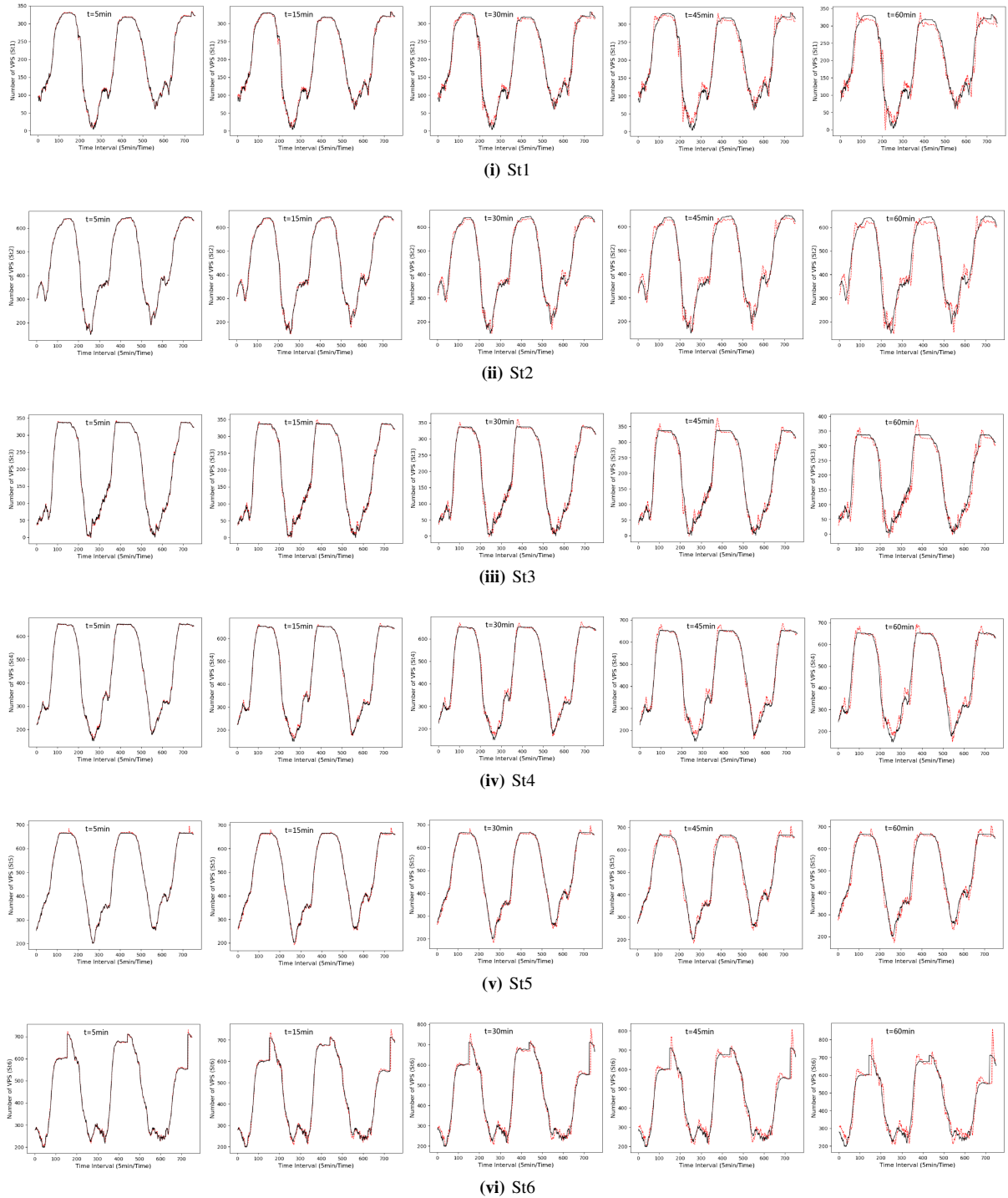

**Fig. 7** Comparisons between the 5-, 15-, 30-, 45- and 60-minute predictions made by ST-GBGRU and actual values (St1-St6)).

and actual values, the better the prediction performance of the model.

The experimental results can be analyzed from the following three aspects. First of all, no matter which one of the ST-GBGRU, GCN+GRU, ConvLSTM-DCN models is used for the experiment, the performance of direct prediction method is always better than that of iterative prediction method. Fig. 9 shows a clear comparison of the gap in prediction performance between the single-step direct prediction and multi-step iterative prediction. The histogram shows the number of tasks for which the direct prediction method is more accurate than the iterative method under three evaluation metrics (i.e., RMSE, MAPE, and MAE), in all 32 prediction tasks (i.e., 15- /30-/45-/60-minute prediction for 8 parking lot). This is because in the process of iterative prediction, the predicted value is used to replace the actual value, resulting in the accumulation of

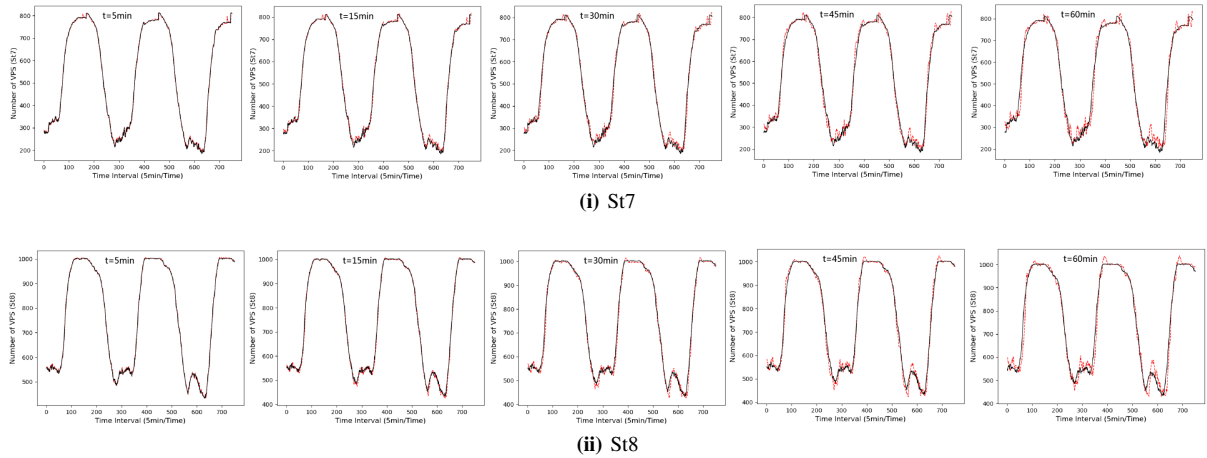

**Fig. 8** Comparisons between the 5-, 15-, 30-, 45- and 60-minute predictions made by ST-GBGRU and actual values (St7-St8).

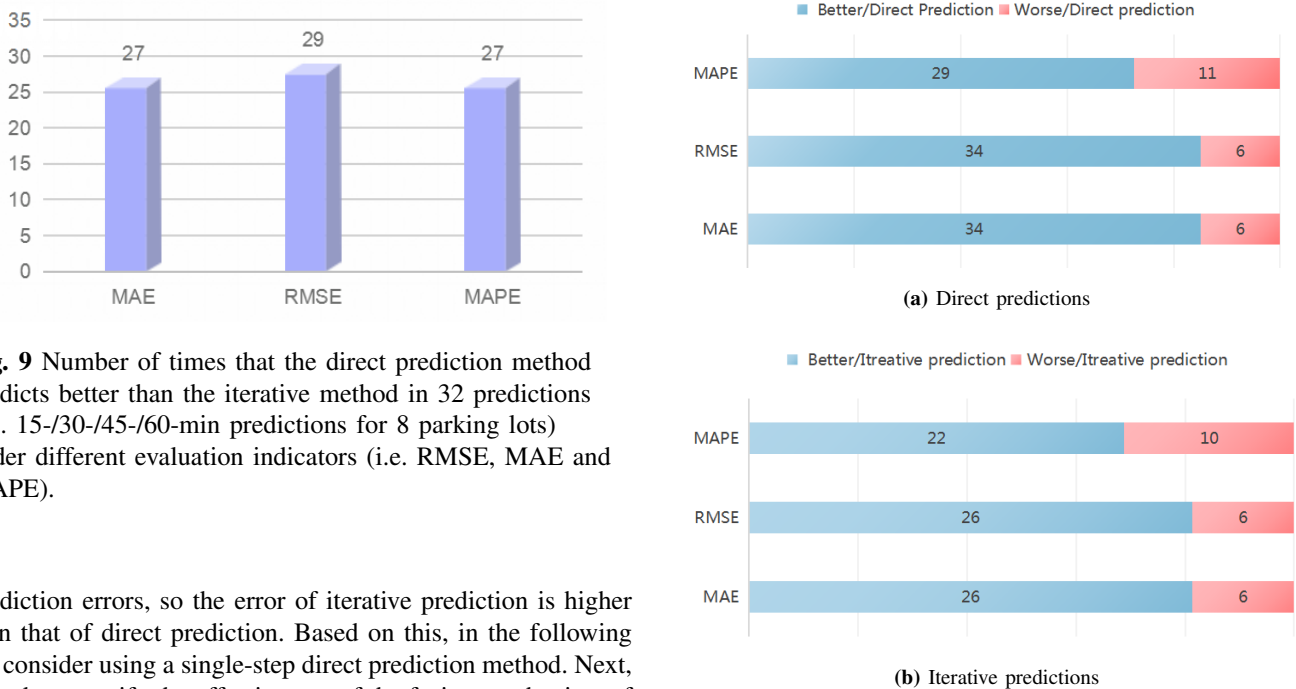

**Fig. 9** Number of times that the direct prediction method predicts better than the iterative method in 32 predictions (i.e. 15-/30-/45-/60-min predictions for 8 parking lots) under different evaluation indicators (i.e. RMSE, MAE and MAPE).

**(a)** Direct predictions

**(b)** Iterative predictions

**Fig. 10** Number of time that the predictions with ST-GBGRU model do better /worse than ConvLSTM-DCN model under different evaluation indicators.

prediction errors, so the error of iterative prediction is higher than that of direct prediction. Based on this, in the following we consider using a single-step direct prediction method. Next, in order to verify the effectiveness of the fusion mechanism of the proposed model, we also compare the performance of the ST-GBGRU model with the simple stacked GCN+GRU model. It can be seen in Table II that the performance of the ST-GBGRU model is always outperforms that of the GCN+GRU model, whether direct prediction or iterative prediction is used. Last but not least, it can be seen from the Fig. 7 and 8 that with the increase of time step, the prediction performance of the model is also declining. Because the prediction is based on time correlation, and with the increase of prediction time steps, the time correlation becomes weaker and weaker. And as shown in Fig. 10, for RMSE, MAE and MAPE, the direct predictions with ST-GBGRU model predicted better on 34, 34 and 29 out of 40 tasks, respectively, than those with ConvLSTM-DCN model, and in iterative predictions, predictions with ST-GBGRU model outperformed those with ConvLSTM-DCN model 26, 26 and 28 times (totally 32 tasks), respectively.

The performance of ST-GBGRU model was also compared with that of LSTM (Long Short-Term Memory), GRU-NN (Gated Recurrent Unit Neural Network), SAE (Stacked Auto-Encoder), SVR (Support Vector Regression), BPNN (Back Propagation Neural Network) and KNN (K-Nearest Neighbor). Grid search is used to determine the hyperparameter settings of these comparison models, which can be seen [15]. However, these models are all for the prediction of a single parking lot. In order to compare the performance, parking lot St7 is selected as the analysis target. Table IV and Fig. 11

**TABLE III Comparisons between the real and predicted numbers of VPSs of St7 (from 14:30 to 15:30, 06/05/2018).**

| Time Point | Real Values | ST-GBGRU model | | | | | ConvLSTM-DCN model | | | | |
|---|---|---|---|---|---|---|---|---|---|---|---|
| | | 5min | 15min | 30min | 45min | 60min | 5min | 15min | 30min | 45min | 60min |
| 16:25 | 278 | 283 | 292 | 283 | 288 | 327 | 282 | 298 | 286 | 302 | 328 |
| 16:30 | 276 | 281 | 285 | 289 | 294 | 325 | 278 | 295 | 292 | 307 | 327 |
| 16:35 | 277 | 277 | 287 | 293 | 299 | 320 | 281 | 293 | 296 | 300 | 325 |
| 16:40 | 280 | 279 | 286 | 294 | 300 | 307 | 279 | 290 | 296 | 294 | 321 |
| 16:45 | 277 | 283 | 281 | 286 | 292 | 315 | 277 | 287 | 287 | 301 | 323 |
| 16:50 | 278 | 282 | 282 | 288 | 294 | 306 | 277 | 285 | 292 | 304 | 319 |
| 16:55 | 284 | 281 | 287 | 286 | 291 | 295 | 279 | 287 | 289 | 305 | 313 |
| 17:00 | 293 | 286 | 286 | 281 | 286 | 302 | 278 | 291 | 281 | 299 | 315 |
| 17:05 | 306 | 294 | 286 | 282 | 286 | 307 | 278 | 293 | 282 | 301 | 318 |
| 17:10 | 323 | 309 | 291 | 287 | 291 | 307 | 283 | 296 | 287 | 299 | 318 |
| 17:15 | 322 | 322 | 300 | 287 | 291 | 297 | 291 | 303 | 286 | 288 | 313 |
| 17:20 | 314 | 325 | 316 | 287 | 292 | 300 | 305 | 315 | 290 | 286 | 313 |
| 17:25 | 321 | 317 | 330 | 292 | 298 | 298 | 321 | 329 | 299 | 293 | 310 |
| MAE | | **5.54** | **10.92** | **17.85** | **18.69** | **25.62** | 10.77 | 11.92 | 18.62 | 22.15 | 28.15 |
| MAPE(%) | | **1.85** | **3.62** | **5.88** | **6.27** | **8.92** | 3.48 | 4.05 | 6.19 | 7.53 | 9.96 |
| RMSE | | **7.01** | **13.86** | **20.81** | **20.16** | **29.63** | 14.15 | 16.90 | 20.83 | 23.72 | 33.49 |

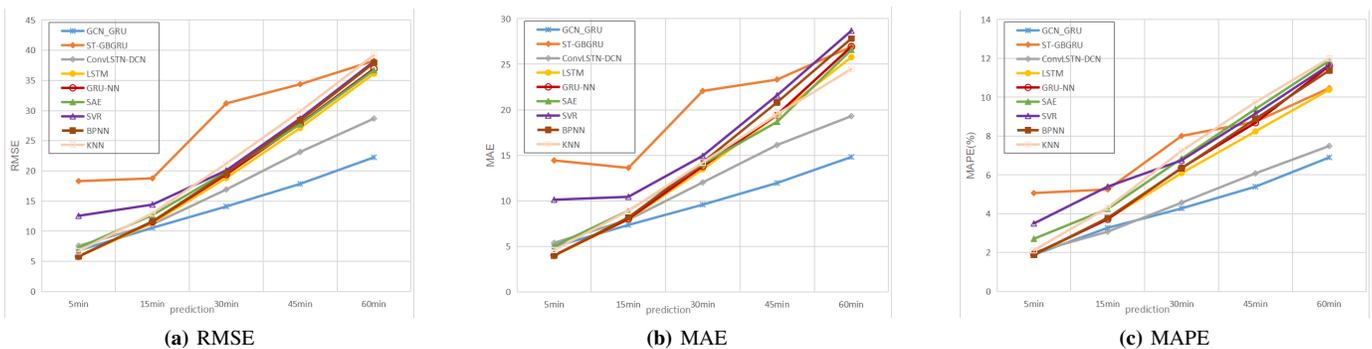

**(a) RMSE**  **(b) MAE**  **(c) MAPE**

**Fig. 11** Comparisons of the predictions of the number of VPSs in St7 between ST-GBGRU and comparison models under different evaluation indicators (i.e. RMSE, MAE and MAPE).

**TABLE IV Comparisons of prediction errors.**

| Models | Indicators | 5min | 15min | 30min | 45min | 60min |
|---|---|---|---|---|---|---|
| ST-GBGRU | RMSE | 6.79 | **10.56** | **14.10** | **17.84** | **22.24** |
| | MAE | 4.94 | **7.38** | **9.58** | **11.96** | **14.81** |
| | MAPE(%) | 1.89 | 3.28 | **4.27** | **5.39** | **6.90** |
| GCN+GRU | RMSE | 18.30 | 18.76 | 31.18 | 34.37 | 38.26 |
| | MAE | 14.43 | 13.61 | 22.06 | 23.31 | 26.92 |
| | MAPE(%) | 5.07 | 5.25 | 8.01 | 8.74 | 10.47 |
| ConvLSTM-DCN | RMSE | 7.57 | 11.14 | 16.91 | 23.14 | 28.67 |
| | MAE | 5.38 | 7.92 | 12.02 | 16.13 | 19.30 |
| | MAPE(%) | 2.01 | **3.08** | 4.57 | 6.08 | 7.49 |
| LSTM | RMSE | **5.77** | 11.35 | 18.82 | 27.11 | 36.13 |
| | MAE | **3.96** | 7.96 | 13.47 | 19.38 | 25.78 |
| | MAPE(%) | **1.87** | 3.74 | 6.09 | 8.24 | 10.39 |
| GRU-NN | RMSE | 5.83 | 11.57 | 19.40 | 27.85 | 36.92 |
| | MAE | 4.01 | 8.02 | 13.79 | 19.45 | 26.93 |
| | MAPE(%) | 1.92 | 3.72 | 6.36 | 8.69 | 11.61 |
| SAE | RMSE | 7.24 | 12.67 | 19.84 | 27.73 | 36.70 |
| | MAE | 5.02 | 8.92 | 14.04 | 18.67 | 26.58 |
| | MAPE(%) | 2.71 | 4.25 | 6.84 | 9.39 | 11.88 |
| SVR | RMSE | 12.55 | 14.41 | 20.08 | 28.69 | 38.14 |
| | MAE | 10.12 | 10.44 | 14.94 | 21.58 | 28.68 |
| | MAPE(%) | 3.50 | 5.40 | 6.75 | 9.15 | 11.66 |
| BPNN | RMSE | 5.81 | 11.63 | 19.62 | 28.37 | 37.91 |
| | MAE | 3.98 | 8.17 | 14.22 | 20.81 | 27.82 |
| | MAPE(%) | 1.88 | 3.79 | 6.34 | 8.88 | 11.38 |
| KNN | RMSE | 6.66 | 12.99 | 21.19 | 29.87 | 39.20 |
| | MAE | 4.59 | 8.90 | 14.21 | 19.50 | 24.47 |
| | MAPE(%) | 2.12 | 4.32 | 7.24 | 9.74 | 12.00 |

show the comparisons results on prediction errors. It can be observed that although the prediction accuracy of the ST-GBGRU in 5 min is slightly lower than that of the LSTM model in 5-min prediction (i.e.MAPE, with a sight difference of 0.02%), our method outperforms all the other prediction tasks in 15, 30, 45, and 60 minutes prediction tasks. Moreover, with the increase of the prediction time step, the prediction advantage of the ST-GBGRU model is greater. Specifically, compared with the ConvLSTM-DCN model in [27], the RMSE of dConvLSTM-DCN framework in 5-/15-/30-/45-/60-min predictions are reduced by 0.77, 0.58, 2.81 and 2.59 and 6.42, respectively.

To verify that the predicted results were statistically significant, p-values were calculated by T-Test. The results are shown in Table V. It can be seen from Table that, except for the 5-minute prediction, the p-values at other times are all less than 0.05, and the results are statistically significant.

We also compare the training time and prediction time of the proposed model and other comparision models (i.e. GCN+GRU, ConvLSTM-DCN [27], dDconvLSTM-DCN [28]) with and without GPU. In terms of training time, the average value of 30 experiments is taken as the final result. And in terms of prediction time, the average value of 10000 independent predictions is taken as the final result. The results in Table VI show that the model proposed in this paper has a great performance improvement in training time and prediction time. Compared with the dConvLSTM-DCN and ConvLSTM-DCN models with complex internal structures, ST-GBGRU can achieve relatively short training time and prediction time due to its simple internal structure. The prpposed ST-GBGRU model can be used in practical applications because of its good prediction performance and short training time, and has great application prospects. And in the case of using GPU, the training time of the model proposed in this paper is only 56.77% of that without using GPU. Although the training speed has been improved, the improvement is not very obvious. Henceforth, considering the cost of GPU and training time, we suggest the use of CPU due to the low cost by using CPU.

**TABLE V P-values between ST-GBGRU and comparison models calculated by T-Test.**

| ComparisonModels | P-Values | | | | |
|---|---|---|---|---|---|
| | 5min | 15min | 30min | 45min | 60min |
| ConvLSTM-DCN | 8.28E-02 | 2.06E-42 | 1.70E-22 | 1.94E-15 | 8.06E-10 |
| GCN+GRU | 4.42E-60 | 1.48E-31 | 4.43E-08 | 2.77E-02 | 3.44E-14 |
| LSTM | 9.79E-21 | 3.74E-94 | 4.27E-10 | 1.46E-34 | 1.46E-34 |
| GRU | 2.49E-38 | 7.35E-16 | 3.89E-02 | 1.82E-04 | 2.24E-04 |
| SAE | 3.40E-15 | 2.93E-07 | 2.99E-15 | 6.90E-05 | 2.85E-03 |
| SVR | 1.02E-03 | 3.82E-08 | 1.71E-10 | 4.22E-02 | 1.69E-09 |
| BPNN | 3.90E-29 | 7.23E-13 | 2.47E-14 | 4.70E-02 | 2.96E-07 |
| KNN | 9.88E-48 | 5.89E-07 | 1.91E-12 | 1.19E-02 | 4.76E-02 |

**TABLE VI Runtime Comparisons among different models.**

| Models | Training time (seconds/epoch) | Prediction time ($\mu s$/Prediction) |
|---|---|---|
| ST-GBGRU(GPU) | 3.23 | 0.53 |
| ST-GBGRU(CPU) | 5.69 | 1.20 |
| dConvLSTM-DCN(CPU) | 254.89 | 15.05 |
| dConvLSTM-DCN(GPU) | 47.56 | 3.86 |
| ConvLSTM-DCN(GPU) | 28.36 | 1.62 |
| LSTM(GPU) | 1.19 | 0.93 |

## III. CONCLUSION

In this paper, we propose a novel model, ST-GBGRU, to predict the number of VPSs, which models both time dependence and space dependence, since parking lots in real scenes do not present regular distribution. Temporal correlation is modeled by GRU and spatial correlation is captured by GCN. The parking lots in Santa Monica are constructed as a graph based on geographical distance. Combined with ST-GBGRU model and two multi-step prediction methods, the number of VPSs were predicted in 5, 15, 30, 45 and 60 minutes. The performance of ST-GBGRU model is compared with other models. The results show that the prediction accuracy of ST-GBGRU model is better than that of ConvLSTM-DCN model, and it can be used to predict the number of VPSs in real scenes.

## ACKNOWLEDGMENT

This work was supported by the Fundamental Scientific Research Project of Wenzhou City under Grant G20190021.